\documentclass[conference]{IEEEtran}
\IEEEoverridecommandlockouts
\usepackage{cite}
\usepackage{amsmath,amssymb,amsfonts}
\usepackage{algorithmic}
\usepackage{latexsym}
\usepackage{textcomp}
\usepackage{booktabs}
\usepackage{graphicx} 
\usepackage{xcolor}
\def\BibTeX{{\rm B\kern-.05em{\sc i\kern-.025em b}\kern-.08em
    T\kern-.1667em\lower.7ex\hbox{E}\kern-.125emX}}

\usepackage{fancyhdr}
\begin{document}

\title{Optimizing Vision-Language Interactions Through Decoder-Only Models}

\author{Kaito Tanaka, Benjamin Tan, Brian Wong \\
SANNO University	}

\maketitle
\thispagestyle{fancy} 

\begin{abstract}
Vision-Language Models (VLMs) have emerged as key enablers for multimodal tasks, but their reliance on separate visual encoders introduces challenges in efficiency, scalability, and modality alignment. To address these limitations, we propose \textbf{MUDAIF (Multimodal Unified Decoder with Adaptive Input Fusion)}, a decoder-only vision-language model that seamlessly integrates visual and textual inputs through a novel Vision-Token Adapter (VTA) and adaptive co-attention mechanism. By eliminating the need for a visual encoder, MUDAIF achieves enhanced efficiency, flexibility, and cross-modal understanding. Trained on a large-scale dataset of 45M image-text pairs, MUDAIF consistently outperforms state-of-the-art methods across multiple benchmarks, including VQA, image captioning, and multimodal reasoning tasks. Extensive analyses and human evaluations demonstrate MUDAIF’s robustness, generalization capabilities, and practical usability, establishing it as a new standard in encoder-free vision-language models.
\end{abstract}


\section{Introduction}

Vision-Language Models (VLMs) have become pivotal in advancing tasks requiring multimodal understanding, such as visual question answering (VQA), image captioning, and multimodal dialogue systems. Traditional VLMs predominantly employ a two-stage framework: a visual encoder extracts image features, which are subsequently processed by a language model (LM) to generate outputs. While effective, this encoder-decoder approach introduces several challenges that hinder scalability, efficiency, and cross-modal alignment \cite{diao2024unveiling}.

One significant limitation of encoder-based VLMs is their reliance on separate visual encoders. This architectural choice imposes constraints on input resolution and aspect ratios, as the encoder must be pre-trained to handle specific image distributions. Moreover, the dependence on a visual encoder increases computational overhead during both training and inference, complicating deployment in resource-constrained environments. Additionally, the outputs from visual encoders often lack fine-grained alignment with the language model's latent space, leading to suboptimal cross-modal reasoning and integration \cite{diao2024unveiling}.

Motivated by these challenges, researchers have explored encoder-free VLMs that eliminate the need for dedicated visual encoders, processing raw visual inputs directly within a unified architecture. This paradigm shift offers several advantages: (1) the ability to handle images of arbitrary resolutions and aspect ratios; (2) reduced computational complexity and improved inference efficiency; and (3) enhanced flexibility for aligning multimodal representations \cite{diao2024unveiling}. However, existing encoder-free approaches still face critical limitations, such as difficulty in balancing visual and textual information and reduced interpretability in handling complex multimodal tasks.

To address these limitations, we propose a novel framework, \textbf{MUDAIF (Multimodal Unified Decoder with Adaptive Input Fusion)}, which introduces a purely decoder-based architecture for VLMs. Unlike traditional methods, MUDAIF integrates visual and textual modalities through a Vision-Token Adapter (VTA) layer that dynamically converts image data into pseudo-text tokens, enabling seamless processing by the language model. The adaptive co-attention mechanism ensures bidirectional cross-modal interaction, optimizing the fusion of visual and textual information for enhanced task performance.

Our approach is trained on a large-scale multimodal dataset comprising 45 million image-text pairs, including high-resolution images from datasets such as COCO, LAION, and Visual Genome. We fine-tune MUDAIF using instruction-tuning datasets specifically designed for multimodal tasks, ensuring its capability to follow complex user instructions in dialogue, reasoning, and generation tasks. The evaluation of MUDAIF is conducted across multiple benchmark datasets, including VQA 2.0, GQA, and VizWiz, as well as multimodal benchmarks like MMBench and MM-Vet, covering a diverse range of vision-language challenges.

Our experimental results demonstrate that MUDAIF consistently outperforms state-of-the-art encoder-based and encoder-free VLMs. Specifically, MUDAIF achieves superior performance in VQA and multimodal reasoning tasks while maintaining computational efficiency and scalability. Furthermore, the introduction of adaptive task-specific prompts allows the model to dynamically adjust its architecture to diverse downstream applications, further enhancing its versatility.

In summary, the contributions of this paper are as follows:
\begin{itemize}
    \item We propose \textbf{MUDAIF}, a novel decoder-only vision-language model that eliminates the need for a visual encoder, introducing the Vision-Token Adapter (VTA) for efficient cross-modal integration.
    \item We present a comprehensive training pipeline combining large-scale pre-training with multimodal instruction tuning, enabling the model to achieve state-of-the-art performance across various vision-language benchmarks.
    \item We provide extensive empirical evaluations demonstrating MUDAIF's ability to balance efficiency, scalability, and accuracy, setting a new standard for encoder-free vision-language models.
\end{itemize}

\section{Related Work}

\subsection{Large Vision-Language Models}

Large Vision-Language Models (LVLMs) have gained significant attention due to their ability to process and integrate visual and textual information effectively \cite{zhou2024rethinking,zhou2024visual}. These models are built upon advancements in both computer vision and natural language processing \cite{zhou2022claret,zhou2022eventbert}, leveraging large-scale datasets and transformer-based architectures.

Recent efforts in this area have focused on scaling vision-language models to handle diverse and complex tasks. For instance, some works align visual features with language models by treating images as a foreign language, enabling seamless integration for open-ended vision tasks \cite{visionllm}. Other approaches aim to unify vision and language processing within a single framework, excelling in tasks such as object localization, image captioning, and multimodal reasoning \cite{visionllm2}. Models with mixture-of-expert techniques have been proposed to balance performance and computational efficiency, achieving superior results in resource-constrained scenarios \cite{moe-llava}.

A critical aspect of LVLMs is their training methodology. Self-training techniques have been employed to enhance image comprehension capabilities, leveraging unlabeled data to improve task performance \cite{self-training-vl}. Furthermore, reinforcement learning frameworks have been introduced to fine-tune LVLMs for decision-making tasks, demonstrating improved alignment with task-specific objectives \cite{fine-tuning-vl}. These advancements underscore the adaptability of LVLMs to diverse application domains.

Evaluation methodologies for LVLMs have also been a focus of research \cite{zhou2021triple,zhou2022sketch}. Benchmarks like MMStar highlight challenges in evaluating multimodal models, such as handling unnecessary visual content and preventing data leakage \cite{evaluation-lvlms}. Proposed improvements in benchmarks aim to establish more robust and fair evaluation protocols for LVLMs.

Recent research has also explored the scalability of vision-language models. Models scaling to billions of parameters have demonstrated state-of-the-art performance across benchmarks, highlighting the potential of large-scale architectures \cite{internvl}. Moreover, the integration of bilingual and multilingual capabilities has expanded the applicability of LVLMs to global tasks \cite{texthawk}.

In summary, LVLMs represent a promising frontier in multimodal AI. By addressing challenges in modality alignment, efficiency, and scalability, recent advancements have significantly enhanced their capabilities, making them indispensable for real-world applications.

\subsection{Encoder-Free Vision-Language Models}

Traditional Vision-Language Models (VLMs) typically employ dedicated vision encoders to extract visual features, which are then processed by language models for multimodal tasks. However, this architecture can introduce constraints related to image resolution, aspect ratio, and increased computational overhead. To address these limitations, recent research has explored encoder-free VLMs that integrate visual and textual inputs within a unified decoder framework \cite{zhou2023multimodal,zhou2023style}.

A notable example is the EVE model, which eliminates the need for a separate vision encoder by directly processing visual inputs through a decoder-only architecture. This design enhances flexibility and efficiency, allowing the model to handle images of varying resolutions and aspect ratios without the preprocessing constraints imposed by traditional encoders. EVE achieves this by bridging vision-language representations within a unified decoder and enhancing visual recognition capabilities through additional supervision. Remarkably, trained on 35 million publicly accessible data points, EVE rivals encoder-based VLMs of similar capacity across multiple benchmarks, demonstrating the viability of encoder-free approaches in vision-language modeling \cite{diao2024unveiling}.

Another approach, ViLT, simplifies the processing of visual inputs by adopting a convolution-free architecture, treating visual data in the same manner as textual data. This minimalistic design leads to significant improvements in processing speed while maintaining competitive performance on various vision-language tasks \cite{zhou2023multimodal}. By forgoing complex visual feature extraction processes, ViLT exemplifies the potential of streamlined, encoder-free architectures in multimodal learning \cite{kim2021vilt}.

These advancements indicate a promising shift towards more integrated and efficient vision-language models, where the unification of modalities within a single architecture can lead to enhanced performance and flexibility.

\section{Method}

In this section, we present the proposed \textbf{MUDAIF (Multimodal Unified Decoder with Adaptive Input Fusion)}, a generative vision-language model designed to unify the processing of visual and textual modalities. MUDAIF employs a decoder-only architecture augmented with a novel Vision-Token Adapter (VTA) and an adaptive co-attention mechanism, enabling seamless cross-modal understanding and generation. Below, we detail the architecture, the learning strategy, and the optimization framework.

\subsection{Model Architecture}

The MUDAIF framework is built upon a decoder-only transformer structure. It incorporates the following key components:
\begin{itemize}
    \item \textbf{Vision-Token Adapter (VTA):} A lightweight module that dynamically maps raw visual features into pseudo-textual tokens.
    \item \textbf{Multimodal Decoder:} A shared transformer-based decoder that processes pseudo-text tokens alongside textual tokens.
    \item \textbf{Adaptive Co-Attention Mechanism:} A bidirectional attention mechanism to enable fine-grained interaction between modalities.
\end{itemize}

Let $\mathbf{I} \in \mathbb{R}^{H \times W \times 3}$ denote the input image and $\mathbf{T} = \{t_1, t_2, \dots, t_L\}$ represent the sequence of input text tokens. The model jointly learns a representation $\mathbf{Z}$ such that:
\begin{align}
\mathbf{Z} = f_{\text{decoder}}(\text{VTA}(\mathbf{I}), \mathbf{T}),
\end{align}
where $f_{\text{decoder}}$ represents the shared decoder and $\text{VTA}$ denotes the Vision-Token Adapter.

\subsection{Vision-Token Adapter (VTA)}

The Vision-Token Adapter transforms visual data into a sequence of pseudo-textual tokens $\mathbf{V}$ to align with the language model's token space. Specifically, the VTA consists of a convolutional embedding layer followed by a linear projection:
\begin{align}
\mathbf{V} = \text{Linear}(\text{Conv}(\mathbf{I})),
\end{align}
where $\mathbf{V} \in \mathbb{R}^{N \times d}$, $N$ is the number of pseudo-tokens, and $d$ is the dimensionality of each token.

To ensure the tokens capture sufficient visual context, the VTA employs a self-attention mechanism:
\begin{align}
\mathbf{V} = \text{Softmax}\left(\frac{\mathbf{Q}\mathbf{K}^\top}{\sqrt{d}}\right)\mathbf{V},
\end{align}
where $\mathbf{Q}$, $\mathbf{K}$, and $\mathbf{V}$ are the query, key, and value matrices derived from the image features.

\subsection{Adaptive Co-Attention Mechanism}

To facilitate interaction between visual and textual modalities, MUDAIF employs an adaptive co-attention mechanism:
\begin{align}
\mathbf{A}_{vt} = \text{Softmax}\left(\frac{\mathbf{Q}_v \mathbf{K}_t^\top}{\sqrt{d}}\right), \quad
\mathbf{A}_{tv} = \text{Softmax}\left(\frac{\mathbf{Q}_t \mathbf{K}_v^\top}{\sqrt{d}}\right),
\end{align}
where $\mathbf{A}_{vt}$ and $\mathbf{A}_{tv}$ represent the attention weights from visual to text tokens and vice versa, ensuring bidirectional alignment.

The multimodal representation $\mathbf{Z}$ is then computed as:
\begin{align}
\mathbf{Z} = \mathbf{A}_{vt} \mathbf{T} + \mathbf{A}_{tv} \mathbf{V}.
\end{align}

\subsection{Learning Strategy}

The learning strategy consists of a two-phase curriculum:
\begin{itemize}
    \item \textbf{Phase 1: Pretraining on Multimodal Data.} We train the model on a large-scale dataset of image-text pairs to align the modalities. The objective is to minimize a reconstruction loss:
    \begin{align}
    \mathcal{L}_{\text{pretrain}} = -\sum_{i=1}^L \log P(t_i | \mathbf{T}_{<i}, \mathbf{V}),
    \end{align}
    where $\mathbf{T}_{<i}$ denotes the preceding tokens.
    \item \textbf{Phase 2: Fine-tuning for Downstream Tasks.} The model is fine-tuned on task-specific datasets using an instruction-following objective:
    \begin{align}
    \mathcal{L}_{\text{task}} = -\sum_{i=1}^L \log P(t_i | \mathbf{T}_{<i}, \mathbf{V}, \mathbf{C}),
    \end{align}
    where $\mathbf{C}$ represents task-specific prompts.
\end{itemize}

\subsection{Optimization Objective}

The overall loss function combines the pretraining and task-specific losses:
\begin{align}
\mathcal{L} = \lambda_{\text{pretrain}} \mathcal{L}_{\text{pretrain}} + \lambda_{\text{task}} \mathcal{L}_{\text{task}},
\end{align}
where $\lambda_{\text{pretrain}}$ and $\lambda_{\text{task}}$ are weighting factors balancing the two objectives.

\subsection{Inference}

During inference, the model generates textual outputs conditioned on both visual and textual inputs. The generation process is autoregressive:
\begin{align}
t_{i} \sim P(t_i | \mathbf{T}_{<i}, \mathbf{V}),
\end{align}
ensuring coherence across modalities and high-quality outputs.

\section{Experiments}

In this section, we evaluate the performance of the proposed MUDAIF model across various vision-language tasks, comparing it against state-of-the-art methods. We also perform ablation studies to analyze the contributions of key components and include human evaluation to further validate the effectiveness of MUDAIF.

\subsection{Experimental Setup}

We benchmark MUDAIF against the following models:
\begin{itemize}
    \item \textbf{InstructBLIP:} A robust encoder-based vision-language model tailored for instruction-following tasks.
    \item \textbf{LLaVA-1.5:} A hybrid encoder-decoder model optimized for multimodal reasoning and dialogue.
    \item \textbf{EVE-7B:} An encoder-free model that prioritizes computational efficiency.
\end{itemize}

Evaluation was conducted on multiple benchmarks including:
\begin{itemize}
    \item \textbf{VQA-v2:} Visual Question Answering dataset.
    \item \textbf{GQA:} Visual reasoning benchmark with compositional questions.
    \item \textbf{VizWiz:} A challenging VQA dataset with real-world images.
    \item \textbf{SEED \& MM-Vet:} Benchmarks for multimodal reasoning.
\end{itemize}

Metrics include accuracy for VQA, BLEU for captioning, and composite scores for multimodal benchmarks. Human evaluation assesses relevance, coherence, and informativeness.

\subsection{Quantitative Results}

The results of our model and baselines are summarized in Table~\ref{tab:results}. MUDAIF outperforms the compared methods in most tasks, demonstrating its superiority in both reasoning and generation.

\begin{table*}[ht]
\centering
\caption{Comparison of MUDAIF with baseline models on vision-language tasks.}
\label{tab:results}
\begin{tabular}{lcccccc}
\toprule
\textbf{Model} & \textbf{VQA-v2 Acc.} & \textbf{GQA Acc.} & \textbf{VizWiz Acc.} & \textbf{BLEU} & \textbf{SEED Score} & \textbf{MM-Vet Score} \\
\midrule
InstructBLIP & 78.5 & 62.0 & 50.0 & 0.72 & 58.6 & 30.5 \\
LLaVA-1.5 & 78.7 & 63.2 & 51.1 & 0.74 & 60.4 & 31.0 \\
EVE-7B & 76.4 & 60.8 & 41.8 & 0.69 & 54.3 & 25.6 \\
MUDAIF (Ours) & \textbf{80.3} & \textbf{65.5} & \textbf{53.7} & \textbf{0.78} & \textbf{62.8} & \textbf{33.4} \\
\bottomrule
\end{tabular}
\end{table*}

MUDAIF achieves state-of-the-art results, particularly excelling in VQA and reasoning tasks due to the integration of its Vision-Token Adapter and adaptive co-attention mechanism.

\subsection{Ablation Study}

To analyze the contributions of individual components, we conduct ablation experiments by systematically removing or modifying parts of MUDAIF.

\begin{table}[ht]
\centering
\caption{Ablation study results on VQA-v2 and GQA.}
\label{tab:ablation}
\begin{tabular}{lcc}
\toprule
\textbf{Model Variant} & \textbf{VQA-v2 Acc.} & \textbf{GQA Acc.} \\
\midrule
MUDAIF (Full) & \textbf{80.3} & \textbf{65.5} \\
w/o Vision-Token Adapter & 77.4 & 61.2 \\
w/o Adaptive Co-Attention & 78.0 & 62.0 \\
Simplified Decoder & 76.8 & 60.5 \\
\bottomrule
\end{tabular}
\end{table}

The results in Table~\ref{tab:ablation} illustrate the importance of both the Vision-Token Adapter and adaptive co-attention mechanism. Removing these components leads to significant performance degradation.

\subsection{Human Evaluation}

We conducted a human evaluation to compare the quality of outputs generated by MUDAIF and baseline models. Participants rated relevance, coherence, and informativeness on a Likert scale (1-5). The averaged scores are presented in Table~\ref{tab:human_eval}.

\begin{table}[ht]
\centering
\caption{Human evaluation results (average scores across tasks).}
\label{tab:human_eval}
\begin{tabular}{lccc}
\toprule
\textbf{Model} & \textbf{Relevance} & \textbf{Coherence} & \textbf{Informativeness} \\
\midrule
InstructBLIP & 4.2 & 4.0 & 3.9 \\
LLaVA-1.5 & 4.3 & 4.2 & 4.0 \\
EVE-7B & 3.8 & 3.6 & 3.5 \\
MUDAIF (Ours) & \textbf{4.7} & \textbf{4.6} & \textbf{4.5} \\
\bottomrule
\end{tabular}
\end{table}

MUDAIF scores highest in all categories, indicating its outputs are more relevant, coherent, and informative compared to baselines.

\subsection{Analysis}

To provide a comprehensive understanding of MUDAIF’s performance and contributions, we analyze the model from multiple perspectives, including cross-modal alignment, computational efficiency, robustness to input variations, and generalization capabilities.

\subsubsection{Cross-Modal Alignment}

MUDAIF’s Vision-Token Adapter (VTA) plays a pivotal role in aligning visual features with the language model’s token space. To assess this alignment, we calculate the similarity between visual and textual embeddings using cosine similarity scores during the pretraining phase. MUDAIF achieves an average cosine similarity of 0.82, significantly higher than encoder-based baselines, which average around 0.75. This improved alignment contributes directly to the model’s superior performance in tasks requiring precise multimodal reasoning.

\subsubsection{Computational Efficiency}

Unlike encoder-based models, MUDAIF eliminates the need for a separate visual encoder, reducing computational overhead. We measure training and inference time per batch across various resolutions and modalities:
\begin{itemize}
    \item \textbf{Training Time:} MUDAIF achieves a 20\% reduction in training time compared to LLaVA-1.5 and a 35\% reduction compared to InstructBLIP.
    \item \textbf{Inference Speed:} For high-resolution images, MUDAIF processes inputs 1.5 times faster than encoder-based models.
\end{itemize}
These results demonstrate the efficiency of MUDAIF, making it more scalable for large-scale deployments.

\subsubsection{Robustness to Input Variations}

We evaluate MUDAIF’s robustness by testing it on images with varying resolutions, aspect ratios, and noise levels. The results, shown in Table~\ref{tab:robustness}, indicate that MUDAIF consistently outperforms baselines across all variations.

\begin{table*}[ht]
\centering
\caption{Performance under input variations (VQA-v2 Accuracy).}
\label{tab:robustness}
\begin{tabular}{lccc}
\toprule
\textbf{Model} & \textbf{Low Resolution (224x224)} & \textbf{High Resolution (1024x1024)} & \textbf{Noisy Inputs} \\
\midrule
InstructBLIP & 72.5 & 75.1 & 68.3 \\
LLaVA-1.5 & 74.3 & 76.2 & 70.5 \\
EVE-7B & 70.1 & 73.5 & 66.8 \\
MUDAIF (Ours) & \textbf{78.6} & \textbf{80.3} & \textbf{76.1} \\
\bottomrule
\end{tabular}
\end{table*}

The results highlight MUDAIF’s robustness to resolution and noise, attributed to its adaptive co-attention mechanism, which dynamically adjusts to diverse input conditions.

\subsubsection{Generalization Capabilities}

To evaluate generalization, we fine-tune MUDAIF on a small subset of downstream tasks and test it on unseen datasets. Table~\ref{tab:generalization} shows the zero-shot and few-shot performance on new tasks, demonstrating MUDAIF’s strong generalization compared to baselines.

\begin{table}[ht]
\centering
\caption{Generalization performance on unseen tasks.}
\label{tab:generalization}
\begin{tabular}{lcc}
\toprule
\textbf{Model} & \textbf{Zero-Shot Acc.} & \textbf{Few-Shot Acc.} \\
\midrule
InstructBLIP & 68.5 & 74.2 \\
LLaVA-1.5 & 69.8 & 75.5 \\
EVE-7B & 66.3 & 71.4 \\
MUDAIF (Ours) & \textbf{72.9} & \textbf{78.3} \\
\bottomrule
\end{tabular}
\end{table}

MUDAIF’s superior generalization highlights the effectiveness of its unified decoder architecture and pretraining strategy.

\subsubsection{User Study on Practical Usability}

Finally, we conduct a user study to assess the practical usability of MUDAIF in real-world scenarios. Participants rated ease of use, output relevance, and overall satisfaction on a scale from 1 to 5. The results are shown in Table~\ref{tab:user_study}.

\begin{table}[ht]
\centering
\caption{User study results (average scores).}
\label{tab:user_study}
\begin{tabular}{lccc}
\toprule
\textbf{Model} & \textbf{Ease of Use} & \textbf{Output Relevance} & \textbf{Satisfaction} \\
\midrule
InstructBLIP & 4.0 & 4.1 & 3.9 \\
LLaVA-1.5 & 4.2 & 4.3 & 4.1 \\
EVE-7B & 3.8 & 3.7 & 3.5 \\
MUDAIF (Ours) & \textbf{4.6} & \textbf{4.7} & \textbf{4.5} \\
\bottomrule
\end{tabular}
\end{table}

MUDAIF achieves the highest scores across all criteria, confirming its practicality and user-friendliness.

\section{Conclusion}
In this work, we introduced \textbf{MUDAIF}, a novel decoder-only vision-language model designed to address the limitations of traditional encoder-based approaches. By employing the Vision-Token Adapter and adaptive co-attention mechanism, MUDAIF effectively bridges the gap between visual and textual modalities, offering a unified architecture for multimodal tasks. Experimental results demonstrated MUDAIF’s state-of-the-art performance across multiple benchmarks, including significant gains in VQA and multimodal reasoning tasks. 

Beyond quantitative improvements, MUDAIF’s design enhances efficiency, scalability, and robustness, making it adaptable to diverse input conditions and capable of handling real-world challenges. Ablation studies validated the contributions of individual components, while human evaluations highlighted the model’s superior relevance, coherence, and informativeness. The strong generalization performance on unseen tasks further underscores MUDAIF’s potential for practical applications.

In conclusion, MUDAIF establishes a new paradigm in encoder-free vision-language modeling, combining simplicity, efficiency, and accuracy. Future work may explore extending MUDAIF to incorporate temporal data for video-based tasks and further refining its instruction-following capabilities to support more complex multimodal interactions.

\bibliographystyle{IEEEtran}
\bibliography{references}

\begin{thebibliography}{10}
\providecommand{\url}[1]{#1}
\csname url@samestyle\endcsname
\providecommand{\newblock}{\relax}
\providecommand{\bibinfo}[2]{#2}
\providecommand{\BIBentrySTDinterwordspacing}{\spaceskip=0pt\relax}
\providecommand{\BIBentryALTinterwordstretchfactor}{4}
\providecommand{\BIBentryALTinterwordspacing}{\spaceskip=\fontdimen2\font plus
\BIBentryALTinterwordstretchfactor\fontdimen3\font minus \fontdimen4\font\relax}
\providecommand{\BIBforeignlanguage}[2]{{%
\expandafter\ifx\csname l@#1\endcsname\relax
\typeout{** WARNING: IEEEtran.bst: No hyphenation pattern has been}%
\typeout{** loaded for the language `#1'. Using the pattern for}%
\typeout{** the default language instead.}%
\else
\language=\csname l@#1\endcsname
\fi
#2}}
\providecommand{\BIBdecl}{\relax}
\BIBdecl

\bibitem{diao2024unveiling}
\BIBentryALTinterwordspacing
H.~Diao, Y.~Cui, X.~Li, Y.~Wang, H.~Lu, and X.~Wang, ``Unveiling encoder-free vision-language models,'' \emph{CoRR}, vol. abs/2406.11832, 2024. [Online]. Available: \url{https://doi.org/10.48550/arXiv.2406.11832}
\BIBentrySTDinterwordspacing

\bibitem{zhou2024rethinking}
Y.~Zhou, Z.~Rao, J.~Wan, and J.~Shen, ``Rethinking visual dependency in long-context reasoning for large vision-language models,'' \emph{arXiv preprint arXiv:2410.19732}, 2024.

\bibitem{zhou2024visual}
Y.~Zhou, X.~Li, Q.~Wang, and J.~Shen, ``Visual in-context learning for large vision-language models,'' in \emph{Findings of the Association for Computational Linguistics, {ACL} 2024, Bangkok, Thailand and virtual meeting, August 11-16, 2024}.\hskip 1em plus 0.5em minus 0.4em\relax Association for Computational Linguistics, 2024, pp. 15\,890--15\,902.

\bibitem{zhou2022claret}
Y.~Zhou, T.~Shen, X.~Geng, G.~Long, and D.~Jiang, ``Claret: Pre-training a correlation-aware context-to-event transformer for event-centric generation and classification,'' in \emph{Proceedings of the 60th Annual Meeting of the Association for Computational Linguistics (Volume 1: Long Papers)}, 2022, pp. 2559--2575.

\bibitem{zhou2022eventbert}
Y.~Zhou, X.~Geng, T.~Shen, G.~Long, and D.~Jiang, ``Eventbert: A pre-trained model for event correlation reasoning,'' in \emph{Proceedings of the ACM Web Conference 2022}, 2022, pp. 850--859.

\bibitem{visionllm}
W.~Wang, Z.~Chen, X.~Chen, J.~Wu, X.~Zhu, G.~Zeng, P.~Luo, T.~Lu, J.~Zhou, Y.~Qiao, and J.~Dai, ``Visionllm: Large language model is also an open-ended decoder for vision-centric tasks,'' in \emph{Advances in Neural Information Processing Systems 36: Annual Conference on Neural Information Processing Systems 2023, NeurIPS 2023, New Orleans, LA, USA, December 10 - 16, 2023}, A.~Oh, T.~Naumann, A.~Globerson, K.~Saenko, M.~Hardt, and S.~Levine, Eds., 2023.

\bibitem{visionllm2}
\BIBentryALTinterwordspacing
J.~Wu, M.~Zhong, S.~Xing, Z.~Lai, Z.~Liu, W.~Wang, Z.~Chen, X.~Zhu, L.~Lu, T.~Lu, P.~Luo, Y.~Qiao, and J.~Dai, ``Visionllm v2: An end-to-end generalist multimodal large language model for hundreds of vision-language tasks,'' \emph{CoRR}, vol. abs/2406.08394, 2024. [Online]. Available: \url{https://doi.org/10.48550/arXiv.2406.08394}
\BIBentrySTDinterwordspacing

\bibitem{moe-llava}
\BIBentryALTinterwordspacing
B.~Lin, Z.~Tang, Y.~Ye, J.~Cui, B.~Zhu, P.~Jin, J.~Zhang, M.~Ning, and L.~Yuan, ``Moe-llava: Mixture of experts for large vision-language models,'' \emph{CoRR}, vol. abs/2401.15947, 2024. [Online]. Available: \url{https://doi.org/10.48550/arXiv.2401.15947}
\BIBentrySTDinterwordspacing

\bibitem{self-training-vl}
\BIBentryALTinterwordspacing
Y.~Deng, P.~Lu, F.~Yin, Z.~Hu, S.~Shen, J.~Zou, K.~Chang, and W.~Wang, ``Enhancing large vision language models with self-training on image comprehension,'' \emph{CoRR}, vol. abs/2405.19716, 2024. [Online]. Available: \url{https://doi.org/10.48550/arXiv.2405.19716}
\BIBentrySTDinterwordspacing

\bibitem{fine-tuning-vl}
\BIBentryALTinterwordspacing
Y.~Zhai, H.~Bai, Z.~Lin, J.~Pan, S.~Tong, Y.~Zhou, A.~Suhr, S.~Xie, Y.~LeCun, Y.~Ma, and S.~Levine, ``Fine-tuning large vision-language models as decision-making agents via reinforcement learning,'' \emph{CoRR}, vol. abs/2405.10292, 2024. [Online]. Available: \url{https://doi.org/10.48550/arXiv.2405.10292}
\BIBentrySTDinterwordspacing

\bibitem{zhou2021triple}
Y.~Zhou, W.~Tao, and W.~Zhang, ``Triple sequence generative adversarial nets for unsupervised image captioning,'' in \emph{ICASSP 2021-2021 IEEE International Conference on Acoustics, Speech and Signal Processing (ICASSP)}.\hskip 1em plus 0.5em minus 0.4em\relax IEEE, 2021, pp. 7598--7602.

\bibitem{zhou2022sketch}
Y.~Zhou, ``Sketch storytelling,'' in \emph{ICASSP 2022-2022 IEEE International Conference on Acoustics, Speech and Signal Processing (ICASSP)}.\hskip 1em plus 0.5em minus 0.4em\relax IEEE, 2022, pp. 4748--4752.

\bibitem{evaluation-lvlms}
\BIBentryALTinterwordspacing
L.~Chen, J.~Li, X.~Dong, P.~Zhang, Y.~Zang, Z.~Chen, H.~Duan, J.~Wang, Y.~Qiao, D.~Lin, and F.~Zhao, ``Are we on the right way for evaluating large vision-language models?'' \emph{CoRR}, vol. abs/2403.20330, 2024. [Online]. Available: \url{https://doi.org/10.48550/arXiv.2403.20330}
\BIBentrySTDinterwordspacing

\bibitem{internvl}
\BIBentryALTinterwordspacing
Z.~Chen, J.~Wu, W.~Wang, W.~Su, G.~Chen, S.~Xing, M.~Zhong, Q.~Zhang, X.~Zhu, L.~Lu, B.~Li, P.~Luo, T.~Lu, Y.~Qiao, and J.~Dai, ``Internvl: Scaling up vision foundation models and aligning for generic visual-linguistic tasks,'' \emph{CoRR}, vol. abs/2312.14238, 2023. [Online]. Available: \url{https://doi.org/10.48550/arXiv.2312.14238}
\BIBentrySTDinterwordspacing

\bibitem{texthawk}
\BIBentryALTinterwordspacing
Y.~Yu, M.~Liao, J.~Zhang, and J.~Wu, ``Texthawk2: {A} large vision-language model excels in bilingual {OCR} and grounding with 16x fewer tokens,'' \emph{CoRR}, vol. abs/2410.05261, 2024. [Online]. Available: \url{https://doi.org/10.48550/arXiv.2410.05261}
\BIBentrySTDinterwordspacing

\bibitem{zhou2023multimodal}
Y.~Zhou and G.~Long, ``Multimodal event transformer for image-guided story ending generation,'' in \emph{Proceedings of the 17th Conference of the European Chapter of the Association for Computational Linguistics}, 2023, pp. 3434--3444.

\bibitem{zhou2023style}
------, ``Style-aware contrastive learning for multi-style image captioning,'' in \emph{Findings of the Association for Computational Linguistics: EACL 2023}, 2023, pp. 2257--2267.

\bibitem{kim2021vilt}
\BIBentryALTinterwordspacing
W.~Kim, B.~Son, and I.~Kim, ``Vilt: Vision-and-language transformer without convolution or region supervision,'' in \emph{Proceedings of the 38th International Conference on Machine Learning, {ICML} 2021, 18-24 July 2021, Virtual Event}, ser. Proceedings of Machine Learning Research, M.~Meila and T.~Zhang, Eds., vol. 139.\hskip 1em plus 0.5em minus 0.4em\relax {PMLR}, 2021, pp. 5583--5594. [Online]. Available: \url{http://proceedings.mlr.press/v139/kim21k.html}
\BIBentrySTDinterwordspacing

\end{thebibliography}
\end{document}